# 3DPalsyNet: A Facial Palsy Grading and Motion Recognition Framework using Fully 3D Convolutional Neural Networks

GARY STOREY[1], RICHARD JIANG[2], SHELAGH KEOGH[1], AHMED BOURIDANE[1] and CHANG-TSUN LI[3].

[1] Department of Computer and Information Sciences, Northumbria University, Newcastle Upon Tyne, NE1 8ST, UK
[2] Department of Computing and Communication, Lancaster University, Lancaster, LA1 4WA, UK
[3] School of Information Technology, Deakin University, Geelong, Australia.

Corresponding author: Richard Jiang (e-mail: r.jiang2@lancaster.ac.uk).

The authors would like to thank the financial support from the EPSRC grant (EP/P009727/1).

**ABSTRACT** The capability to perform facial analysis from video sequences has significant potential to positively impact in many areas of life. One such area relates to the medical domain to specifically aid in the diagnosis and rehabilitation of patients with facial palsy. With this application in mind, this paper presents an end-to-end framework, named 3DPalsyNet, for the tasks of mouth motion recognition and facial palsy grading. 3DPalsyNet utilizes a 3D CNN architecture with a ResNet backbone for the prediction of these dynamic tasks. Leveraging transfer learning from a 3D CNNs pre-trained on the Kinetics data set for general action recognition, the model is modified to apply joint supervised learning using center and softmax loss concepts. 3DPalsyNet is evaluated on a test set consisting of individuals with varying ranges of facial palsy and mouth motions and the results have shown an attractive level of classification accuracy in these task of 82% and 86% respectively. The frame duration and the loss function affect was studied in terms of the predictive qualities of the proposed 3DPalsyNet, where it was found shorter frame duration's of 8 performed best for this specific task. Centre loss and softmax have shown improvements in spatio-temporal feature learning than softmax loss alone, this is in agreement with earlier work involving the spatial domain.

## I. INTRODUCTION

THE task of action recognition is a computer vision problem that has been subject to a significant amount of research for varying actions types. Specific sub-tasks within this area have been studied such as human motion, sports and facial exposition recognition where varying degrees of success have been shown [1], [2]. Within the medical domain there is significant interest in technology which can successfully detect human actions, primarily for medical pathologies which affect an individual's neuromuscular system resulting in atypical movements. Through tracking the levels of atypical motion over time clinicians can establish the current severity of both progressive and regressive conditions. One such condition is facial palsy, in which sudden onset in the loss of facial muscle motion occurs due to damage to the cranial nerve. This nerve damage produces an extreme asymmetrical appearance which can be especially significant in the eyes, brow and mouth regions of the face both when at rest and during the forming of facial expressions. Previous medical research [3]–[5] has highlighted the correlation between patient outcomes and the diagnosis and rehabilitation prescribed by trained medical professionals, specialised therapy plans tailored via regular feedback resulted in the best patient outcomes [6]. The potential to use such a system on a smart device has the potential to provide the clinician with more regular objective feedback on the condition and tailor therapy without always needing to physically see the patient. This is especially beneficial in scenarios where the distance between clinician and patient is large or the availability of either party to meet is limited. As the face plays a major role during interpersonal communication and facial expression the onset of facial palsy can have a significant psychological impact upon the patients. The capability to track rehabilitation privately within a comfortable setting like their own home may also provide a benefit to some patients.

To develop an automated system that can assist medical

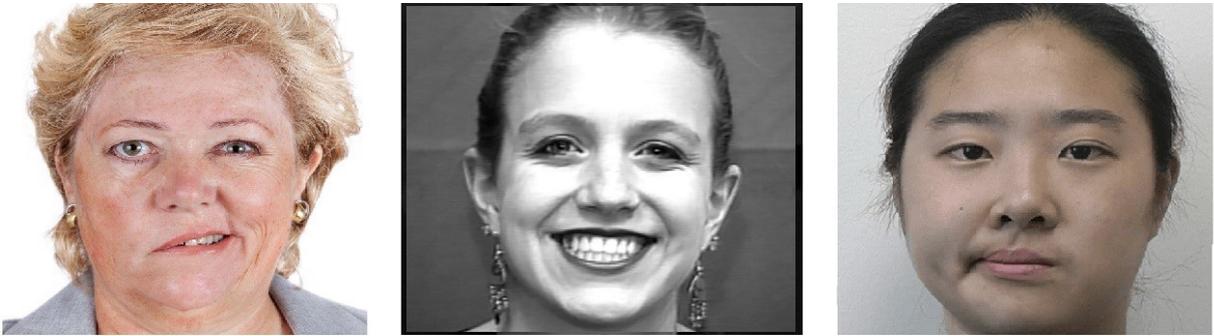

FIGURE 1. When is a smile a smile. The left image show the smile of a facial palsy patient, the centre a smile and on the right a asymmetrical motion which is similar to a smile

professionals in the tracking and planning of a facial palsy rehabilitation plan, there are a number of challenges. Two key functions of a potential system are the capability to recognise facial motions and grade the facial palsy level. In a clinical setting the medical professional would guide the patient through a range of specific facial motions for facial palsy grading so that the medical professional supervision would ensure correct motions were carried out. The challenge of recognising specific facial motions, for example a smile, have been heavily researched especially in facial expression recognition problems; however in the case of facial palsy recognising the asymmetrical nature of the facial motion adds a further challenge [1]. Fig.1 provides an example of this specific challenge. In the case of automated grading of facial palsy only a small amount of research has been conducted mainly limited to traditional methods using Local Binary Pattern features and a Support Vector Machine for classifier on a small sample size [7], [8]. Recently, [9] have applied a Convolutional Neural Network (CNN) based method to a larger data set of 2000 images. While the method has shown promising results the technique still uses images rather than video data. It is known that the temporal information available from video data can provide further discriminative information to ascertain a facial expression [10]. This temporal information also has the potential to boost facial palsy grading thus providing the capability to examine the range of motion across an entire action rather than a single frame of the current methods.

The recognition tasks from video sequences are still challenging and as such they have yet to show the dramatic increase in performance accuracy that has occurred in detection tasks from static images. While approaches applying deep learning based methods have been proposed, such as Recurrent Neural Networks (RNN) [11], Two-stream [12] and C3D [13], from the research to date each method has shown some limitations. RNN based networks have been shown to be incapable of capturing the powerful convolutional features for recognition tasks [1]. Two-stream methods use both image data and optical flow features to represent the spatial and temporal data, respectively, and have shown to produce some of the most promising results through they require pre-processed optical flow features that adds additional computational overhead. While the C3D method uses 3D convolutional layers to learn spatio-temporal features and has demonstrated good performance accuracy on the sport action data set, it does not generalise well to other more complex recognition tasks [2]. This is mainly due to the relatively small video data sets available for optimising the large number of parameters in 3D CNNs. In addition the C3D network is shallow in comparison to the state-of-the-art architectures used in image based recognition tasks where deeper networks have generally performed better. The introduction of a new Kinetics data set [14] that contains 300,000 videos has provided a large scale data set has the potential to train deep 3D CNNs that have the capability to generalise well to other action recognition tasks [2].

Recently, the research team developed a new multi-task framework for joint face detection and facial landmarks locating, namely Integrated Deep Model (IDM), which has been demonstrated with robust performance on face and landmark detection. Based on this initial work, a further novel framework 3DPalsyNet, for facial palsy diagnosis is proposed, where the IDM is cascaded with two further specific 3D ResNet components that are designed to detect mouth motion and carry out palsy level grading, respectively. Fig. 2 shows the schematic view of the new 3DPalsyNet framework. In the framework, besides engaging the IDM model to address the challenge of facial palsy analysis, a fully 3D end-to-end CNN architecture with ResNet backbone was specifically designed, while the framework leverages the Integrated Deep Model [15] to initially perform face detection on video sequence frames. The fully 3D end-to-end CNN network is then trained via transfer learning for mouth motion estimation and facial palsy grading, respectively.

In summary, the novel contribution to knowledge, outlined in this paper, includes,

1) Extending the IDM to video-based facial modelling and proposing 3DPalsyNet a new framework for facial palsy analysis;

2) 3DPalsyNet a new framework which includes a 3D

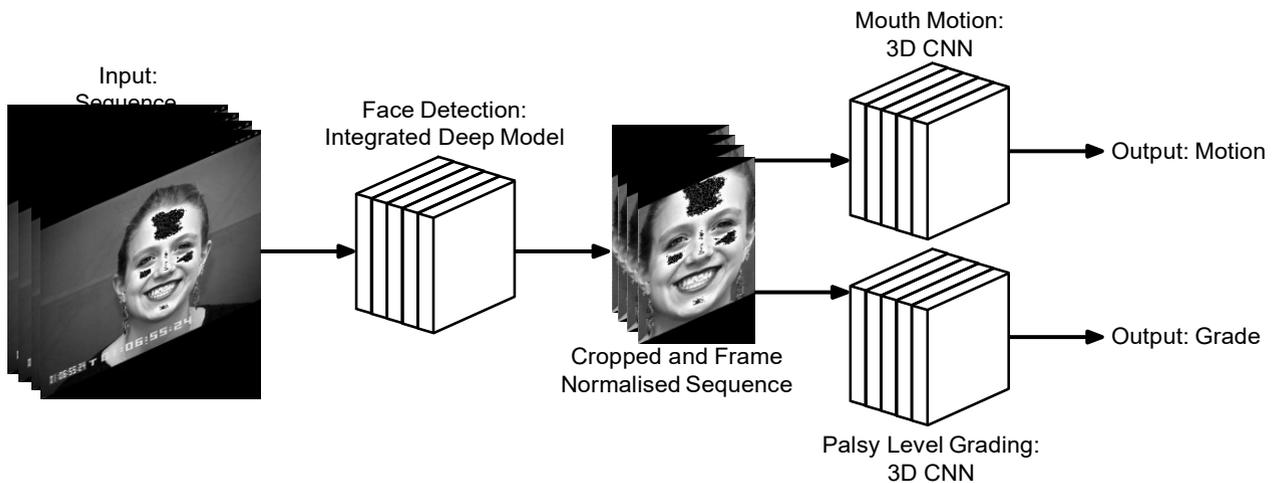

FIGURE 2. Overview of the proposed new 3DPalsyNet framework for facial palsy and mouth motion analysis.

CNN architecture using ResNet backbone to address the needs of facial mouth motion and facial palsy grading, respectively.

3) To train the 3D CNNs, a new Center Loss based transfer learning scheme was developed for the spatio-temporal domain. We also carry out transfer learning via training on the Kinetics dataset, and apply the learned model for the two tasks.

The ablation experiments were designed to investigate the effect of loss function and frame duration on classification accuracy.

The remainder of this paper comprises of a review of relevant work within section II, followed by an in-depth overview of the methods proposed within section III. Section IV is a discussion of the experiments undertaken and the results obtained. Section V presents a conclusion.

## II. RELATED WORK

The task of action recognition is well established within the field of computer vision, with applications ranging from identifying sports based upon the movement of the participants [16] to human facial emotion recognition [17]. Unlike the methods applied in object detection which deals with only the spatial domain, the learning of discriminative temporal domain features from motion data across $n$ frames of a video sequence adds further challenges to action recognition task. A selection of the methods proposed for this challenge are discussed within this section.

### A. CLASSICAL METHODS

Prior to the rise of deep learning and convolutional neural networks many techniques were proposed to extract spatio-temporal features from videos frames for action recognition problems. Optical flow is a well established method that depicts the pattern of apparent motion of image objects between two consecutive frames, caused by the motion of objects. More recently, Liu et al. [18] proposed a new optical flow based feature, called Main Directional Mean Optical-Flow (MDMO), which is a variant of Histogram of Optical Flow (HOOF). This feature was validated on 36 separate regions of interest on the subject's face and has shown to produce a very compact feature vector with each region being described by only two values (the direction and magnitude of the optical flow vector). Optical flow features are represented by a 2D vector where each vector is a displacement vector showing the movement of points from first frame to second. As discussed later in this section optical flow is still useful within some state-of-the-art methods [12]. Local Binary Patterns on Three Orthogonal Planes (LBP-TOP) were proposed for facial texture motion in Zhao et al. [19] and found popularity for action recognition problems due to their ability to describe motion textures efficiently. Further improvement to this method to reduce the feature size were proposed in [20].

### B. 2D CONVOLUTIONAL NEURAL NETWORKS

The two-stream 2D CNN-based approach for action recognition has proven to be a popular techniques with this field. Originally proposed by Simonyan et al. [12] the two-streams refer to one stream which takes RGB images data for computing appearance features and the second stream extracts stacked optical flow features to provide discriminative motion information. The combination of both appearance and motion information resulted in improved results in the benchmark action recognition performance at the time of publication on the UCF-101 [21] and HMDB-51 [22] data sets. The two-stream method has been further studied to improve action recognition performance [23]–[25]. However, the generation of stacked optical flow features usually result in an increased computational complexity to this architecture.

## C. 3D CONVOLUTIONAL NEURAL NETWORKS

Recently, 3D CNN-based approaches have begun to show promise in the task of action recognition as they have been able to leverage the introduction of large-scale training data sets. In contrast to the two-stream methods described previously these architectures require only a single input to the network in the form of a video stacked as a set of individual frames. The extension to 3D convolutional kernels intuitively allows for the shift from the spatial domain to feature domain in the spatio-temporal domain, where the $3^{rd}$ dimension captures the motion across the temporal plane. One of the first fully 3D CNN based models was proposed in Tran et al. which they termed C3D [13]. The model used fully 3D convolutional kernels applying the Sports-1M data set [16] for training of the models parameters. Through model evaluations they found that $3 \times 3 \times 3$ convolutional filters produced the best performances. Expansion of the temporal length showed further improvements in recognition accuracy to the 3CD model were reported in [26]. In the same study it was reported that applying optical flows as inputs to the 3D CNN resulted in a higher level of performance than can be obtained from RGB inputs with the best performance being achieved when using a combination of RGB and optical flows. 3D CNN architectures using the Kinetics data set for training from scratch displayed results that were comparable with the results of ImageNet trained 2D CNN architectures in [14]. Recently, complex 3D CNN architectures have been explored where initial studies were limited to shallow ResNet architectures [27]. However, more recently this has been expanded to much deeper ResNets with up to 152 layers and other architectures including ResNeXt-101 [2] which has shown to achieve the best performance on the Kinetics test set. The study has also found that a Kinetics data set pre-trained on simple 3D architectures outperforms complex 2D CNN architectures both on the UCF-101 and HMDB-51 data sets, respectively.

## III. METHOD

This section discusses the 3DPalsyNet framework proposed for the tasks of mouth motion recognition and facial palsy grading. 3DPalsyNet is comprised of two distinct stages: the initial stage relates to video pre-processing employing face detection and landmark localisation to locate the faces from each frame of the sequence. The Integrated Deep Model is used in this stage. The detected face images are then cropped to the face and the number of frames per sequences are normalised to fixed length. The second stage comprises of two 3D CNNs one for each of the face analysis tasks. The proposed 3DPalsyNet framework is shown in Fig.2.

### A. FACE DETECTION AND VIDEO SEQUENCE PRE-PROCESSING

The Integrated Deep Model (IDM) [15] allows for accurate face detection which has shown to provide both high recall and precision. The requirement for accurate face detection is essential to ensure that faces from each frame of the video sequences are extracted for the second stage of the framework. The IDM method leverages a cascaded approach integrating a Faster R-CNN network trained for face detection and a Facial Alignment Network (FAN) to strengthen face detection precision. This integration is achieved through a heat map transformation and integrates a loss function. Given the heat map output of the FAN as $H = h_1, h_2, ..., h_n$ where each $h_i$ is a $n \times m$ matrix equal in dimensions to the input image for the $i$th facial landmark, each value in $h_i$ corresponds to the probability of the facial landmark being located at that specific pixel location within a given face image. We propose a novel method as given by equation (1) to transform the heat map $H$ to a probability score that can be applied to the task of face detection by integrating it with the loss function of the Faster R-CNN face detector.

$$p_{fan} = \frac{1}{N} \sum_{i=n} \max(H_i)\gamma_i \quad (1)$$

Given by the maximum probability $\max(H_i)$ for the $i^{th}$ facial landmark a specific scaling factor $\gamma_i$ is applied for the corresponding landmark. The sum of the scaled probability is then normalised and can be considered as the probability of a face detection derived from the FAN network defined as $p_{fan}$. The scaling value $\gamma$ is primarily introduced to deal with wide ranging face poses, in which certain landmarks retain visibility across all poses, where others become occluded. Two values are applied to $\gamma$ where facial landmarks that are visible across all facial poses are given a value of $\gamma = 1$, while other landmarks are given $\gamma = 0.75$. The values for $\gamma$ were selected as they have been shown to perform optimally in Storey et al. [15].

$$p_{face} = \frac{(p_{fan} + (p_{faster}\delta))}{2} \quad (2)$$

The next step is to define the joint probability of a face region termed as $p_{face}$ defined in equation (2) where $p_{faster}$ is the probability based upon the output of the trained Faster Face features. The penalisation factor $\delta$ is specifically introduced for situations where extremely small detections are classed in the very high 90% probability range as being faces when they are not. The value of $\delta$ is determined by equation (3) where $det$ is the width of the face detection box and $img$ is the width of the image. It is worth noting that a probability penalisation is only applied when a face width is less that 2% of the total image width. Finally the $p_{face}$ is used within the loss function for the face detection classification as described in equation (4).

$$\delta = \begin{cases} 0.7 & \text{if } det * (100 / img) \leq 2 \\ 1 & \text{otherwise} \end{cases} \quad (3)$$

$$loss_{face} = \frac{1}{N_{cls}} \sum_{i=n} -(1-p_i^*) \cdot log(1-p_{face,i}) \\ -p_i^* \cdot log(p_{face,i}) \quad (4)$$

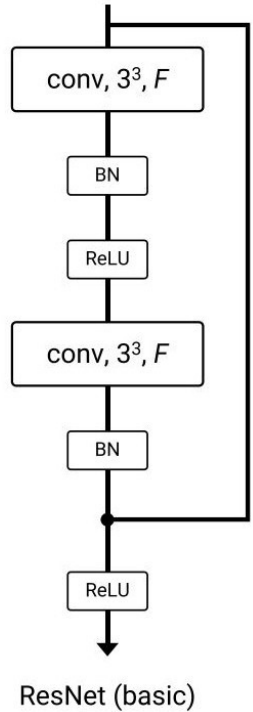

**FIGURE 3.** ResNet Block Architecture.

### B. 3D CNN ARCHITECTURE

3D convolutional architectures are a natural extension of the 2D counterparts that have been widely applied successfully to many image classification tasks. While 2D convolution filters have proven efficacy at learning discriminative features in the spatial domain, they lack the capability to extract spatio-temporal features in action classification tasks, where the input is typically video sequences. The 3rd dimension of the 3D CNN provides the mechanism to learn these spatio-temporal features. The proposed 3D CNN method adopts the ResNet [28] architecture as the backbone of the network, this architecture has been highly successful for image classification tasks. The capacity to develop deep ResNet architectures is related to the use of shortcut connections, allowing the data signal to bypass one layer and moves to the next layer in the sequence; this permits the gradients flow from later layers to the early layers. A basic ResNet block consists of two convolutional layers (Fig. 3 highlights the block design), and each convolutional layer is followed by a batch normalisation and a ReLU. A shortcut pass connects the top of the block to the layer just before the last ReLU in the block.

Unlike previous 3D CNN works in [2] we adopt the joint supervised learning of both softmax loss and center loss. It has been shown in other facial analysis tasks that using softmax loss only results in large intra-class variations of the learned features [29]. Therefore, the adoption and use of center loss will improve the inherent inter-class dispensation and intra-class compactness.

$$loss_{softmax} = \sum_{i=1}^{m} \log \frac{e^{W_{y_i}^T x_i + b_{y_i}}}{\sum_{j=1}^{n} e^{W_j^T x_i + b_j}} \quad (5)$$

Softmax loss is calculated as given in equation (5), where $x_i \in R^d$ is the $i$th feature of the $y_i$th class, the feature dimension is defined by $d$. $W_j \in R^d$ denotes the $j$th column of the weights $W \in R^{dn}$ in the last fully connected layer and $b \in R^n$ is the bias term. Mini-batch size and the total number of class are defined as $m$ and $n$, respectively.

$$loss_{center} = \frac{1}{2} \sum_{i=1}^{m} \| x_i - c_{y_i} \|_2^2 \quad (6)$$

Center loss is defined in equation (6), where $c_{y_i} \in \mathbb{R}^d$ is the $y_i$th class centre of the learnt feature. The feature centers are updated after each mini-batch of training data. The total loss of the network is calculated by equation (7), where $\lambda$ is used for balancing the two loss functions. Center loss is a significantly larger value and therefore requires scaling down. Based upon the experimentation in [29] a value of $\lambda = 0.001$ is used within the proposed 3D CNN.

$$loss_{total} = loss_{softmax} + \lambda loss_{center} \quad (7)$$

### C. 3D CNN MODEL TRAINING

Both of the proposed 3D CNN architectures are trained with the following protocols for their specific tasks. Initially a ResNet18 model is pre-trained on the Kinetics data set for the action recognition task [2]. Transfer learning is then used to train the models for their respective facial analysis task, where the initial layers weight parameters are frozen, only the last convolutional layers parameters and the fully connected layers trained. The layers of the network are trained using a hybrid data set by combining samples from the CK+ emotion and a facial palsy data set with relevant labels for the associated task (Section IV details the breakdown of the dataset and the associated class labelling). To address the class imbalance within the dataset a weighted sampling is employed so that each mini-batch has a similar distribution of class labels. Prior to the training process the video sequences in the training set are first passed through the face detection stage of the 3DPalsyNet and the faces are extracted. The extracted face sequences are then re-sized spatially to 112 pixels x 112 pixels and temporally to $n$ total frames. In this work we consider different values for $n$. When a sequence is less than $n$ frames duplicate frames are interpolated into the sequence while those greater than $n$ have frames removed at equally spaced intervals. Data augmentation techniques are applied to increase the total samples. To help avoid overfitting random flipping, rotation and colour jitter with 50% probability are employed. Two stochastic gradient descent optimisers are then applied to train the network in order to model and fine tune the parameters and to tune the center loss parameters. The training parameters include a learning rate of 0.1, with a weight decay of 0.001 and 0.9 for momentum. Each model was trained for 50 epochs which was sufficient to minimise model loss.

## IV. EXPERIMENTAL EVALUATION

This section presents a thorough experimental evaluation of the proposed 3DPalsyNet framework for both facial palsy grading and mouth motion recognition. All experiments are conducted using PyTorch 0.4 on Windows 10 with a Nvidia GTX 1080 GPU.

For the evaluation of the proposed method data the Extended Cohn-Kanade (CK+) [30] and a Facial Palsy dataset were used. The CK+ database consists of 593 sequences generated from 113 subjects, while the facial palsy data set consists of 696 different sequences with 17 subjects collected from online sources. Since the CK+ sequences range from a neutral face and ends at the full expression, they are aligned with the facial palsy dataset by adding reversed frames, so that the last frame is also a neutral expression. While all samples from the CK+ data set are posed, the facial palsy set contains both posed motions and also general motions such as talking. In the case of mouth motion recognition each sequence is labelled as follows: no motion, smile, mouth open and other mouth motions. For the grading of facial palsy the labelling follows the House-Brackmann scale as shown in table.1, which is commonly applied by medical professionals.

To test the models accuracy for the two classification tasks a leave-one-subject-out (LOSO) protocol is adopted. This is to allow for the testing on unseen faces thus reducing any potential overfitting to previously seen faces. In practice we do not build the models to test all subjects in the data set. 10 subjects have been used for the evaluation process; they are split equally into 5 having facial palsy (Subjects 1 to 5) and 5 who do not have facial palsy (Subjects 6 to 10). The 10 selected subjects cover the total range of labels for both tasks. Therefore, in total there are 397 samples used for the evaluation.

### A. MOUTH MOTION RECOGNITION

Figure 4 provides the overall results for the mouth motion recognition, where it was found that the proposed model has a good predictive capability in this task producing an F1 Score of 82%. In the Figure 6 the results for each of the LOSO test sets are given, we find that all subjects perform reasonably well with F1 Scores close to 80% with the exception of subject 5. On inspection (Figure 5 shows the confusion matrix for this sample) this subject and the samples which prove difficult to classify correctly are an example of an issue which reduce the accuracy for all subjects. This surrounds the overlap in motions that occur between those labelled as others and the rest. There are motions which are similar to a smile, due to the frame normalisation resulting in the possible loss of frames which can differentiate these motions. As this method also uses the global features for learning features it is possible that other motions such as those from the movement of the eyes and brows show overlap across classes therefore reducing accuracy.

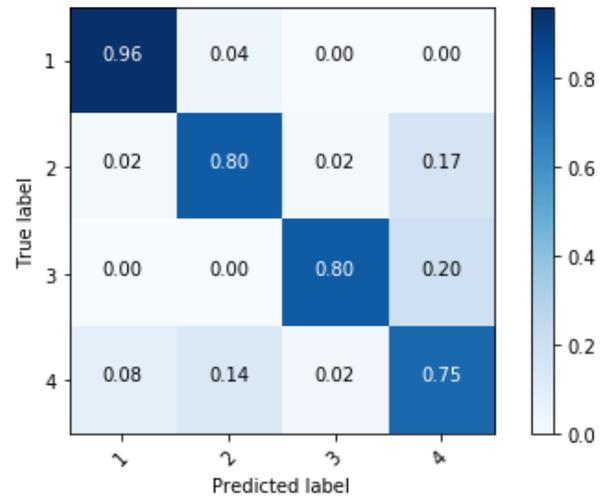

FIGURE 4. Overall Mouth Motion Confusion Matrix.

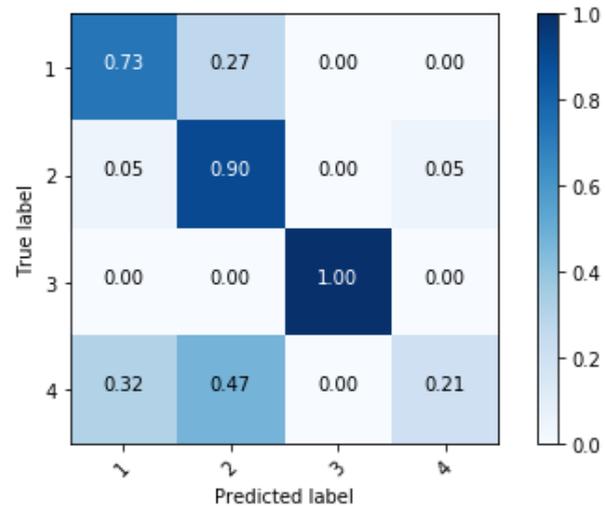

FIGURE 5. Subject 5 Mouth Motion Confusion Matrix.

### B. FACIAL PALSY GRADING

The overall results for the facial palsy grading evaluation are shown in Figure 7. It was found that the proposed model provides a high level of accuracy with a F1 Score of 88%. In Figure 6 the results for each of the LOSO test sets are depicted, showing that all subjects from the CK+ data sets, have a palsy grading label of 1 and are all correctly classified. Subject 1 shows a very poor accuracy in comparison to all other subjects. Subject 1 has 29 samples, out of the 20 incorrect grading 16 are within 1± grades. Subject 1 is a specifically difficult set of sequences as most of the facial sulting in more computational overhead of the method. In action recognition work of [2] a frame duration of 16 were found to work well, as the task of face motion are typically shorter in duration this study proposes to evaluate shorter frame duration. In this experiment the performance effect on the frame duration is evaluated. Table.2 illustrates the F1 scores achieved for each duration over the test sets for frame duration of 8, 12 and 16. From the results it can be seen that a frame duration of 8 seems to give the best performances. It is to be noted that there are samples which are correctly classified in the larger frame duration but incorrectly graded in the 8 frame duration. This is due to the lack of uniformity across motion duration in these tasks. Not only does this parameter have an affect on the accuracy presented by the model, it also has a large effect on the computational overhead of the framework. This can be seen in Table.2 where the use of an additional 4 frames adds about 1 hour to the time the model took to train for 50 epochs of the data set.

| Grade | Impairment |
|---|---|
| 1 | Normal. |
| 2 | Mild dysfunction (slight weakness, normal symmetry at rest). |
| 3 | Moderate dysfunction (obvious but not disfiguring weakness, normal symmetry at rest) Complete eye closure w/ maximal effort, good forehead movement. |
| 4 | Moderately severe dysfunction (obvious and disfiguring asymmetry) Incomplete eye closure, moderate forehead movement. |
| 5 | Severe dysfunction (barely perceptible motion). |
| 6 | Total paralysis (no movement). |

**TABLE 1.** House-Brackmann Facial Paralysis Grading Scale.

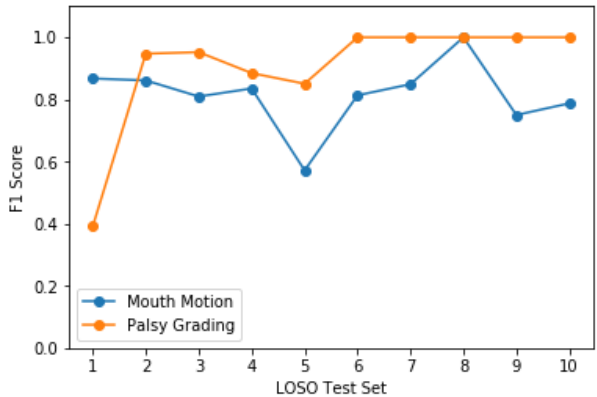

**FIGURE 6.** F1 Score by Subject Test Set.

| Frame Duration | F1 Score | Training Time |
|---|---|---|
| 8 | 86% | 2h 10m |
| 12 | 71% | 3h 20m |
| 16 | 79% | 4h 25m |

**TABLE 2.** Frame Duration Results

expression are not posed but of the individual during normal conversation.

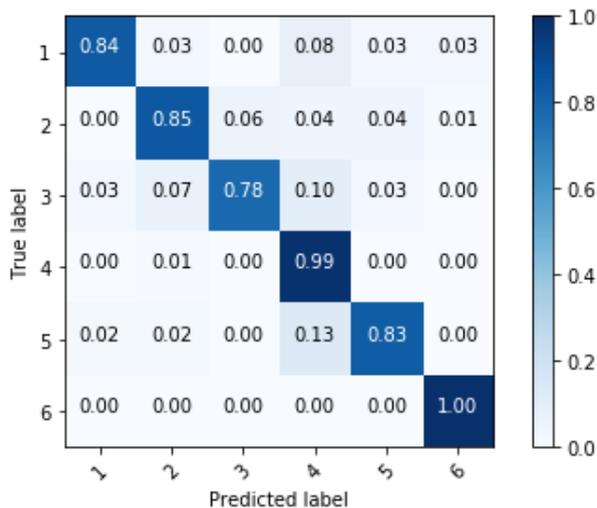

**FIGURE 7.** Overall Palsy Level Grading Results.

### C. ABLATION STUDY - FRAME DURATION

Frame duration is a potentially significant parameter when processing video sequences. Reducing the sequences to a short frame duration can remove important features while long frame duration's may add redundant information re-

### D. ABLATION STUDY - LOSS FUNCTION

A joint supervised method for model training, applying both center loss and softmax loss, has demonstrated the capacity to learn a more discriminative feature representation in the spatial domain, then when applying softmax loss alone. In this paper the experiment has been revisited for the spatio-temporal domain, specifically modified for the proposed 3DPalsyNet framework. The study used Subjects 1 to 5 and the results obtained are shown in Fig.8. For the 366 samples in the facial palsy test, it was found that F1 scores of 86% and 82% for center and softmax loss and softmax loss alone, respectively. This has resulted in a small improvement of the performances as might be expected in image recognition problems. On the other hand, the results obtained for mouth motion recognition have shown to more difficult to improve as demonstrated by a significant decrease of F1 score going from 82% to 49% when also applying center loss.

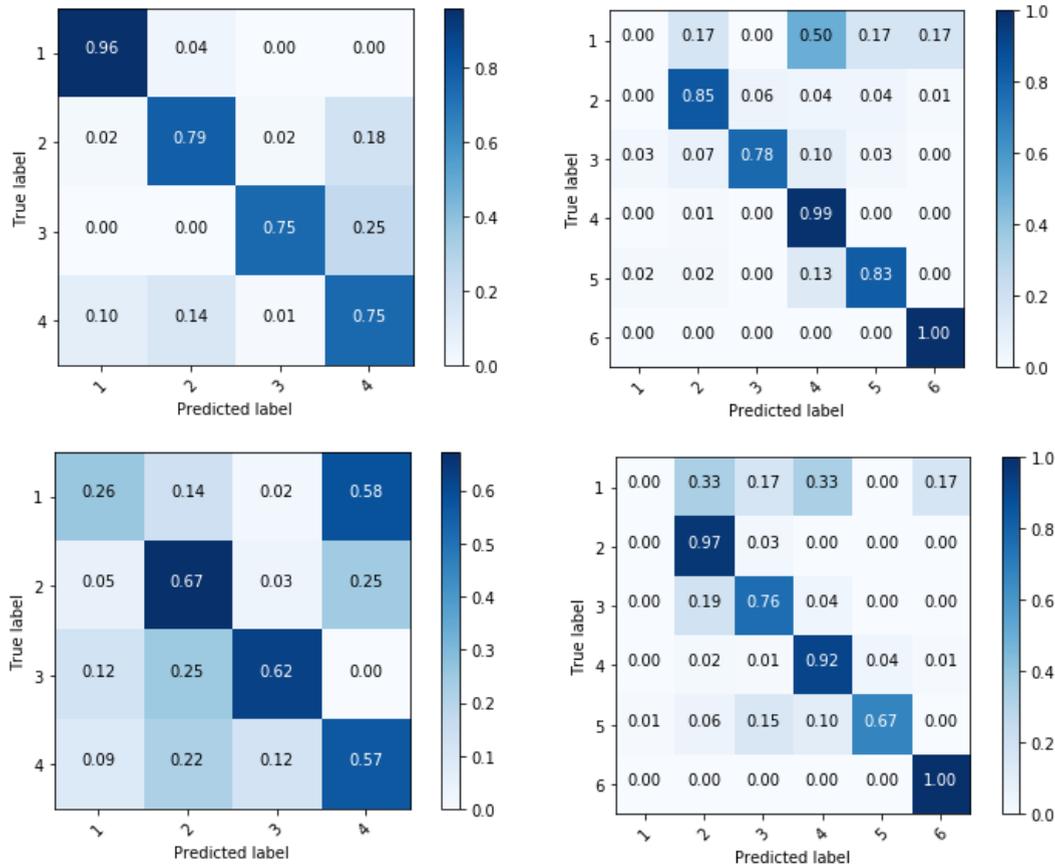

**FIGURE 8.** Loss Function Results. Top Row - 3DPalsyNet Center Loss scheme, Bottom Row - Softmax Loss

## V. CONCLUSIONS

This paper has presented a fully end-to-end framework named 3DPalsyNet for mouth motion detection and facial palsy grading using a modified 3D CNN architecture with an ResNet backbone for capturing the dynamic actions of the video data. The architecture has been evaluated using 2 datasets achieving an F1 score of 82% and 88% F1 score for the mouth motion and facial palsy grading, respectively. The proposed method can be a useful aid for facial palsy grading to assist in the rehabilitation process. It has also been demonstrated that there is potential in using pre-trained Kinetics based 3D CNN for tasks outside of general action detection.

While the results are promising there are many areas in which this research can be taken forward. Firstly there is the potential to investigate more complex backbone and deeper networks, which in this work is limited due to the computational overheads of 3D CNN's. As shown in the frame duration experiment this parameter plays a part in the model accuracy but, there is not universal best parameters when sequences length can vary within the data set; there is potential to look at other pre-processing rather than simple frame duplication or reduction. Specifically for the task presented in this work a larger set of labelling for mouth motion is required to better separate similar facial motions.

### REFERENCES

[1] S. Li and W. Deng, "Deep Facial Expression Recognition: A Survey," 4 2018.
[2] K. Hara, H. Kataoka, and Y. Satoh, "Can Spatiotemporal 3D CNNs Retrace the History of 2D CNNs and ImageNet?" in 2018 IEEE/CVF Conference on Computer Vision and Pattern Recognition. IEEE, 11 2018, pp. 6546–6555.
[3] L. E. Ishii, "Facial Nerve Rehabilitation," Facial Plastic Surgery Clinics of North America, vol. 24, no. 4, pp. 573–575, 11 2016.
[4] P. Guerreschi, P.-E. Gabert, D. Labbé, and V. Martinot-Duquennoy, "Paralysie faciale chez ïA˘Zenfant," Annales de Chirurgie Plastique Esthétique, vol. 61, no. 5, pp. 513–518, 10 2016.
[5] S. Monini, A. Buffoni, M. Romeo, M. Di Traglia, C. Filippi, F. Atturo, and M. Barbara, "Kabat rehabilitation for BellâA˘Zs palsy in the elderly," Acta Oto-Laryngologica, pp. 1–5, 12 2016.
[6] R. W. Lindsay, M. Robinson, and T. A. Hadlock, "Comprehensive Facial Rehabilitation Improves Function in People With Facial Paralysis: A 5-Year Experience at the Massachusetts Eye and Ear Infirmary," Physical Therapy, vol. 90, no. 3, pp. 391–397, 3 2010.
[7] T. Wang, J. Dong, X. Sun, S. Zhang, and S. Wang, "Automatic recognition of facial movement for paralyzed face," Bio-Medical Materials and Engineering, vol. 24, pp. 2751–2760, 2014.
[8] T. Wang, S. Zhang, J. Dong, L. Liu, and H. Yu, "Automatic evaluation of the degree of facial nerve paralysis," Multimedia Tools and Applications, vol. 75, no. 19, pp. 11 893–11 908, 2016.
[9] M. Sajid, T. Shafique, M. Baig, I. Riaz, S. Amin, S. Manzoor, M. Sajid, T. Shafique, M. J. A. Baig, I. Riaz, S. Amin, and S. Manzoor, "Automatic Grading of Palsy Using Asymmetrical Facial Features: A Study Complemented by New Solutions," Symmetry, vol. 10, no. 7, p. 242, 6 2018.

[10] I. Cohen, N. Sebe, A. Garg, L. S. Chen, and T. S. Huang, "Facial expression recognition from video sequences: temporal and static modeling," Computer Vision and Image Understanding, vol. 91, no. 1-2, pp. 160–187, 7 2003.
[11] Z. Yu, G. Liu, Q. Liu, and J. Deng, "Spatio-temporal convolutional features with nested LSTM for facial expression recognition," Neurocomputing, vol. 317, pp. 50–57, 11 2018.
[12] K. Simonyan and A. Zisserman, "Two-Stream Convolutional Networks for Action Recognition in Videos," in Proceedings of the 27th International Conference on Neural Information Processing Systems - Volume 1, 6 2014, pp. 568–576.
[13] D. Tran, J. Ray, Z. Shou, S.-F. Chang, and M. Paluri, "ConvNet Architecture Search for Spatiotemporal Feature Learning," 2017.
[14] W. Kay, J. Carreira, K. Simonyan, B. Zhang, C. Hillier, S. Vijayanarasimhan, F. Viola, T. Green, T. Back, P. Natsev, M. Suleyman, and A. Zisserman, "The Kinetics Human Action Video Dataset," 5 2017.
[15] G. Storey, A. Bouridane, and R. Jiang, "Integrated Deep Model for Face Detection and Landmark Localization From "In The Wild" Images," IEEE Access, vol. 6, 2018.
[16] A. Karpathy, G. Toderici, S. Shetty, T. Leung, R. Sukthankar, and L. Fei-Fei, "Large-Scale Video Classification with Convolutional Neural Networks," in 2014 IEEE Conference on Computer Vision and Pattern Recognition. IEEE, 6 2014, pp. 1725–1732.
[17] Y. Fan, X. Lu, D. Li, and Y. Liu, "Video-based emotion recognition using CNN-RNN and C3D hybrid networks," in Proceedings of the 18th ACM International Conference on Multimodal Interaction - ICMI 2016. New York, New York, USA: ACM Press, 2016, pp. 445–450.
[18] Y.-J. Liu, J.-K. Zhang, W.-J. Yan, S.-J. Wang, G. Zhao, and X. Fu, "A Main Directional Mean Optical Flow Feature for Spontaneous Micro-Expression Recognition," IEEE Transactions on Affective Computing, vol. 3045, no. c, pp. 1–1, 2015.
[19] G. Zhao, M. Pietikäinen, and M. Pietikainen, "Dynamic texture recognition using local binary patterns with an application to facial expressions." IEEE Transactions on Pattern Analysis and Machine Intelligence, vol. 29, no. 6, pp. 915–928, 6 2007.
[20] Y. Wang, J. See, R. C.-W. W. Phan, and Y.-H. H. Oh, "Efficient spatio-temporal local binary patterns for spontaneous facial micro-expression recognition." PloS one, vol. 10, no. 5, p. e0124674, 1 2015.
[21] K. Soomro, A. R. Zamir, and M. Shah, "UCF101: A Dataset of 101 Human Actions Classes From Videos in The Wild," 12 2012.
[22] H. Kuehne, H. Jhuang, E. Garrote, T. Poggio, and T. Serre, "HMDB: A large video database for human motion recognition," in 2011 International Conference on Computer Vision. IEEE, 11 2011, pp. 2556–2563.
[23] C. Feichtenhofer, A. Pinz, and R. P. Wildes, "Spatiotemporal Residual Networks for Video Action Recognition," in Proceedings of the 30th International Conference on Neural Information Processing Systems., 11 2016, pp. 3476–3484.
[24] C. Feichtenhofer, A. Pinz, and A. Zisserman, "Convolutional Two-Stream Network Fusion for Video Action Recognition," in 2016 IEEE Conference on Computer Vision and Pattern Recognition (CVPR). IEEE, 6 2016, pp. 1933–1941.
[25] C. Feichtenhofer, A. Pinz, and R. P. Wildes, "Spatiotemporal Multiplier Networks for Video Action Recognition," in 2017 IEEE Conference on Computer Vision and Pattern Recognition (CVPR). IEEE, 7 2017, pp. 7445–7454.
[26] G. Varol, I. Laptev, and C. Schmid, "Long-Term Temporal Convolutions for Action Recognition," IEEE Transactions on Pattern Analysis and Machine Intelligence, vol. 40, no. 6, pp. 1510–1517, 6 2018.
[27] K. Hara, H. Kataoka, and Y. Satoh, "Learning Spatio-Temporal Features with 3D Residual Networks for Action Recognition," in 2017 IEEE International Conference on Computer Vision Workshops (ICCVW). IEEE, 10 2017, pp. 3154–3160.
[28] K. He, X. Zhang, S. Ren, and J. Sun, "Deep Residual Learning for Image Recognition," in 2016 IEEE Conference on Computer Vision and Pattern Recognition (CVPR). IEEE, 6 2016, pp. 770–778.
[29] Y. Wen, K. Zhang, Z. Li, and Y. Qiao, "A Discriminative Feature Learning Approach for Deep Face Recognition." Springer, Cham, 2016, pp. 499–515.
[30] P. Lucey, J. F. Cohn, T. Kanade, J. Saragih, Z. Ambadar, and I. Matthews, "The Extended Cohn-Kanade Dataset (CK+): A complete dataset for action unit and emotion-specified expression," in 2010 IEEE Computer Society Conference on Computer Vision and Pattern Recognition - Workshops. IEEE, 6 2010, pp. 94–101.

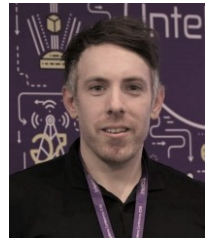

GARY STOREY received his MSc degree in computer science from Northumbria University, Newcastle upon Tyne, United Kingdom in 2015 where he is currently pursuing the PhD degree within the Department of Computer and Information Sciences.

His research interests include computer vision, machine learning and human face analysis and the potential application domains. Awarded best student paper at the Intelligent Systems Conference (IntelliSys) 2018.

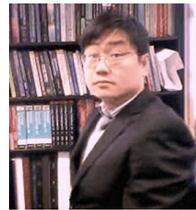

RICHARD JIANG is currently a Senior Lecturer (Associate Professor) in the department of Computing and Communication, Lancaster University, Lancaster, United Kingdom. He was a lecturer in Computer Science in Northumbria University. He is currently leading a team of over 10 research students and postdocs with a focus on machine learning, data security and intelligent systems.

His research interests mainly reside in the fields of Artificial Intelligence, Man-Machine Interaction, Visual Forensics, and Biomedical Image Analysis. His research has been funded by EPSRC, BBSRC, TSB, EU FP, and industry funds, and he has authored and co-authored more than 50 publications.

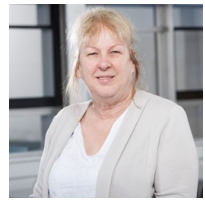

SHELAGH KEOGH was awarded a BSc (Hons) Information Technology, first class, 1999, Sunderland University, UK, and a Master of Research in Informatics from Northumbria University, UK 2006. Currently she is undertaking a PhD study in computer science at Northumbria University. Since 2000 she has been employed as a senior lecturer in Northumbria University. Her research interests include social networks, virtual environments and machine learning for secure networks.

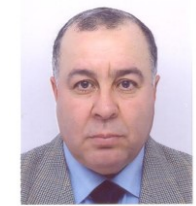

AHMED BOURIDANE (M'98-SM'06) received the "Ingenieur d'État" degree in electronics from Ecole Nationale Polytechnque of Algiers (ENPA), Algeria, in 1982, the M.Phil. degree in electrical engineering (VLSI design for signal processing) from the University of Newcastle-Upon-Tyne, U.K., in 1988, and the Ph.D. degree in electrical engineering (computer vision) from the University of Nottingham, U.K., in 1992.

From 1992 to 1994, he worked as a Research Developer in telesurveillance and access control applications. In 1994, he joined Queen's University Belfast, Belfast, U.K., initially as Lecturer in computer architecture and image processing and then as a Reader in computer science. He became a Professor in Image Engineering and Security at Northumbria University at Newcastle (U.K.) in 2009. His research interests are in imaging for forensics and security, biometrics, homeland security, image/video watermarking and cryptography. He has authored and co-authored more than 200 publications.

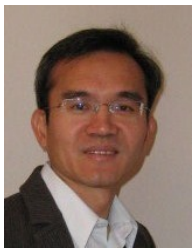

**CHANG-TSUN LI** received the BSc in electrical engineering from National Defence University (NDU), Taiwan, the MSc in computer science from U.S. Naval Postgraduate School, USA and the PhD in computer science from the University of Warwick, UK. He was an associate professor of the Department of Electrical Engineering at NDU during 1998-2002 and a visiting professor of the Department of Computer Science at U.S. Naval Postgraduate School in the second half of 2001. He was a professor of the Department of Computer Science at the University of Warwick (UK) until January 2017 and a professor of Charles Sturt University (Australia) from January 2017 to February 2019. He is currently Professor of Cyber Security of the School of Information Technology at Deakin University, Australia and Research Director of Deakinâ˘A´Zs Centre for Cyber Security Research and innovation.

His research interests include multimedia forensics and security, biometrics, data mining, machine learning, data analytics, computer vision, image processing, pattern recognition, bioinformatics, and content-based image retrieval. The outcomes of his multimedia forensics research have been translated into award-winning commercial products protected by a series of international patents and have been used by a number of police forces and courts of law around the world. He is currently Associate Editor of the EURASIP Journal of Image and Video Processing (JIVP) and Associate of Editor of IET Biometrics. He involved in the organisation of many international conferences and workshops and also served as member of the international program committees for several international conferences. He is also actively contributing keynote speeches and talks at various international events.

...